\documentclass{article}

\usepackage{arxiv}

\usepackage[utf8]{inputenc} 
\usepackage[T1]{fontenc}    
\usepackage{hyperref}       
\usepackage{url}            
\usepackage{amsmath}
\usepackage{booktabs}       
\usepackage{amsfonts}       
\usepackage{caption}
\usepackage{nicefrac}       
\usepackage{microtype}      
\usepackage{lipsum}
\usepackage{graphicx}
\usepackage{float}
\usepackage{tabularx}
\usepackage{pdflscape}

\title{Negative controls reveal volume-driven confounding in radiomics and imaging foundation model features}

\author{
 Katy L. Scott \\
  Princess Margaret Cancer Centre\\
  University Health Network\\
  Toronto, ON \\
   \And
 Sejin Kim \\
  Princess Margaret Cancer Centre\\
  University Health Network\\
  Toronto, ON \\
  \And
 Joshua Siraj \\
  Princess Margaret Cancer Centre\\
  University Health Network\\
  Toronto, ON \\
  \And
  Caryn Geady \\
  Princess Margaret Cancer Centre\\
  University Health Network\\
  Toronto, ON\\
  \And
  Matthew Boccalon \\
  Princess Margaret Cancer Centre\\
  University Health Network\\
  Toronto, ON\\
  \And
  Mattea Welch \\
  Princess Margaret Cancer Centre\\
  University Health Network\\
  Toronto, ON \\
  \And
  Mogtaba Alim \\
  Princess Margaret Cancer Centre\\
  University Health Network\\
  Toronto, ON \\
  \And
  Andrew J. Hope \\
  Radiation Medicine Program \\
  Princess Margaret Cancer Centre\\
  Toronto, ON \\
  \texttt{andrew.hope@uhn.ca} \\
  \And
  Benjamin Haibe-Kains\\
  Princess Margaret Cancer Centre\\
  University Health Network\\
  Toronto, ON \\
  \texttt{benjamin.haibe-kains@uhn.ca} \\
}

\begin{document}
\maketitle
\begin{abstract}
Radiomics and imaging foundation models promise non-invasive biomarkers of tumour biology, yet predictive signatures may reflect tumour volume or acquisition artifacts rather than meaningful image structure. We introduce READII-2-ROQC, an open-source framework that uses volume-preserving negative controls to assess whether radiomic and deep imaging features capture independent spatial signals. READII-2-ROQC generates voxel-perturbed images across tumour, background and whole-image regions using configurable randomization strategies, then compares feature behaviour and model performance between original and control images. Applied to three public cancer imaging cohorts, the framework processed 3,552 tumour volumes and extracted PyRadiomics and foundation-model features from original images and nine matched controls. Reproducing published survival and HPV-status signatures, we show that multiple models retain performance after spatial structure is destroyed, revealing volume-driven or contextual confounding, whereas others show perturbation-sensitive signal. READII-2-ROQC provides a scalable quality-control strategy for developing interpretable, biologically grounded imaging biomarkers and reproducible radiomics workflows.
\end{abstract}


\section{Introduction}
As oncology shifts toward precision medicine, the identification of reliable imaging biomarkers remains essential for non-invasive diagnosis and outcome prediction. Radiological imaging is a cornerstone of modern oncology, integral to diagnosis, disease monitoring, treatment planning, and the development of novel therapeutic protocols. This modality encompasses two-dimensional (e.g., X-ray) and three-dimensional techniques, with computed tomography (CT), magnetic resonance imaging (MRI), and positron emission tomography (PET) being the most prevalent. It is collected routinely during clinical care and offers a non-invasive, cost-effective alternative to high-dimensional data types such as genomics. This clinical utility, coupled with advancements in computational power, artificial intelligence (AI), and data storage, has driven the rapid expansion of radiomics over the last decade \cite{Aerts2014-og,Gillies2016-yt,Lambin2012-oq,Majumder2024-eb,Rogers2020-os}.

Radiomics operates on the hypothesis that the 2D pixel or 3D voxel (volume pixel) intensity distributions in medical images reflect underlying biological processes that can be quantified through high-throughput feature extraction. These quantitative signatures may correlate with genomic profiles, tissue histology, and other critical clinical phenotypes \cite{Majumder2024-eb, Shur2021-vh, Tomaszewski2021-uw}. Radiomic features are typically extracted from a defined region of interest (ROI), like a tumour, identified manually by a radiologist or via automated deep-learning algorithms. Meaningful radiomic features are aggregated into a "radiomic signature" to predict clinical endpoints, including tumour stage, disease progression, overall survival, and treatment response. Standardized features are generally categorized into four classes: shape, intensity-based (first-order), textural (second-order), and higher-order or deep features.

Shape features quantify the geometric and morphological properties of the ROI, such as its volume, surface area, and compactness, independent of voxel intensity. Intensity-based features describe the distribution of voxel values within the ROI via histogram analysis; notably, these exclude spatial information and are highly sensitive to intensity binning strategies \cite{Majumder2024-eb}. Textural features analyze the spatial relationships between voxel intensities within the ROI, typically to quantify heterogeneity \cite{Aerts2014-og}. Lastly, higher-order features utilize convolutional filters (e.g., Wavelet, Gaussian) to uncover complex, non-linear patterns in the data. This may also refer to features extracted using a deep learning method, such as a convolutional neural network (CNN) or foundation model \cite{Pai2024-wc}. 

The proliferation of radiomics has led to an explosion of feature extraction software. While many extraction tools remain proprietary, several open-source platforms have been published for widespread use. PyRadiomics \cite{van-Griethuysen2017-wn}, the Local Image Feature Extraction (LIFEx) \cite{Nioche2018-nt}, and the Foundation Model for Cancer Image Biomarkers (FMCIB) \cite{Pai2024-wc} extractors are the most widely adopted. PyRadiomics has been highly integrated into diverse end-to-end pipelines \cite{Bionic-TMH2023-rw, Pasini2022-dn, Reiazi2021-kb, Shi2019-jz, starmans2025automatedmachinelearningframework}, ranging from traditional clinical workflows to recent agentic AI applications \cite{Tzanis2025-xc}.

Despite the technical maturity of extraction tools, the clinical implementation of radiomics is hindered by concerns regarding interpretability and biological validity \cite{Horvat2024-la, Mu2022-pc}. While numerous studies have linked radiomic signatures to cancer patient survival \cite{Jha2022-su, Vicini2022-kd}, a growing body of evidence suggests that many prognostic and predictive features may actually be surrogates for tumour volume rather than independent biomarkers \cite{Ger2019-nb, Traverso2020-zx, Velichko2021-oq, Welch2019-fv}. A previous study \cite{Welch2019-fv} demonstrated this by reproducing an established four-feature radiomic signature from Aerts et al. \cite{Aerts2014-og}, utilizing the same datasets with one caveat: features were extracted using an original ROI mask but from a modified version of the base image, generated by randomly shuffling the voxel values. The prognostic performance remained unchanged despite the loss of texture signal, confirming that the signature was essentially capturing tumour volume alone. These results have been corroborated by independent studies in head and neck, breast, liver, and lung cancers \cite{Ger2019-nb, Kelahan2022-ox, Traverso2020-zx, Velichko2021-oq}. This distinction is vital for the emerging paradigm of "Radiological-Biological Radiomics Dictionaries," which seek to map quantitative features to specific biological pathways; if a feature is merely a proxy for volume, its biological interpretability and utility in explainable AI frameworks are fundamentally compromised \cite{Gorji2026-vx, Jouzdani2026-ni}.

\begin{figure}[!t]
    \centering
    \includegraphics[width=\textwidth]{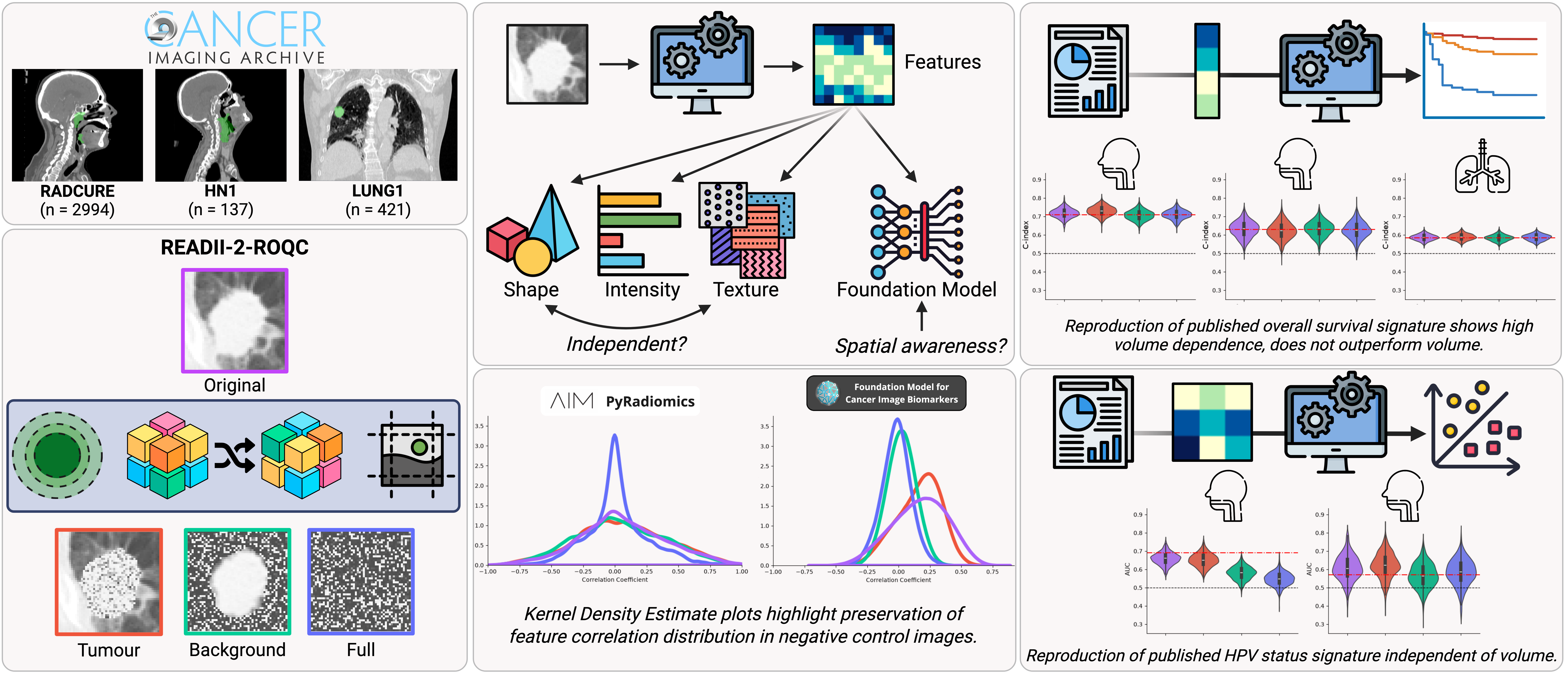}
    \caption{
    Overview of study. Three public imaging datasets were processed through the READII-2-ROQC pipeline and underwent radiomic and deep feature extraction. The relationship between features and geometric qualities of the region of interest was investigated via correlation analyses and the reproduction of published radiomic signatures for overall survival and HPV status prediction.
    }
    \label{fig:overview}
\end{figure}

In this study, we introduce a quality assessment strategy for volume dependence of radiomic signatures in the Radiomic Extraction and Analysis for DICOM Images to Refine Objective Quality Control (READII-2-ROQC) pipeline. This tool generates "negative control" images through voxel permutation to disrupt meaningful texture and spatial signals while preserving the exact volume and shape of the original ROI. We extract radiomic and deep learning features from PyRadiomics and the Foundation Model for Cancer Imaging Biomarkers (FMCIB), respectively, to investigate their relationship to ROI geometry and predictive capabilities across nine negative control configurations (Fig. \ref{fig:overview}). We show that two radiomic survival models retain performance after spatial structure is destroyed, revealing volume-driven or contextual confounding, and that FMCIB features are drawn from the background of a given volume, rather than the tumour. By providing a rigorous framework to decouple biological signal from geometric artifacts, the READII-2-ROQC pipeline serves as a critical bridge between experimental radiomics and clinical implementation, ensuring that the next generation of agentic AI systems and foundation models are powered by truly prognostic, high-fidelity imaging biomarkers.

\section{Results}
\label{sec:results}
\begin{figure}[h]
    \centering
    \includegraphics[width=\textwidth]{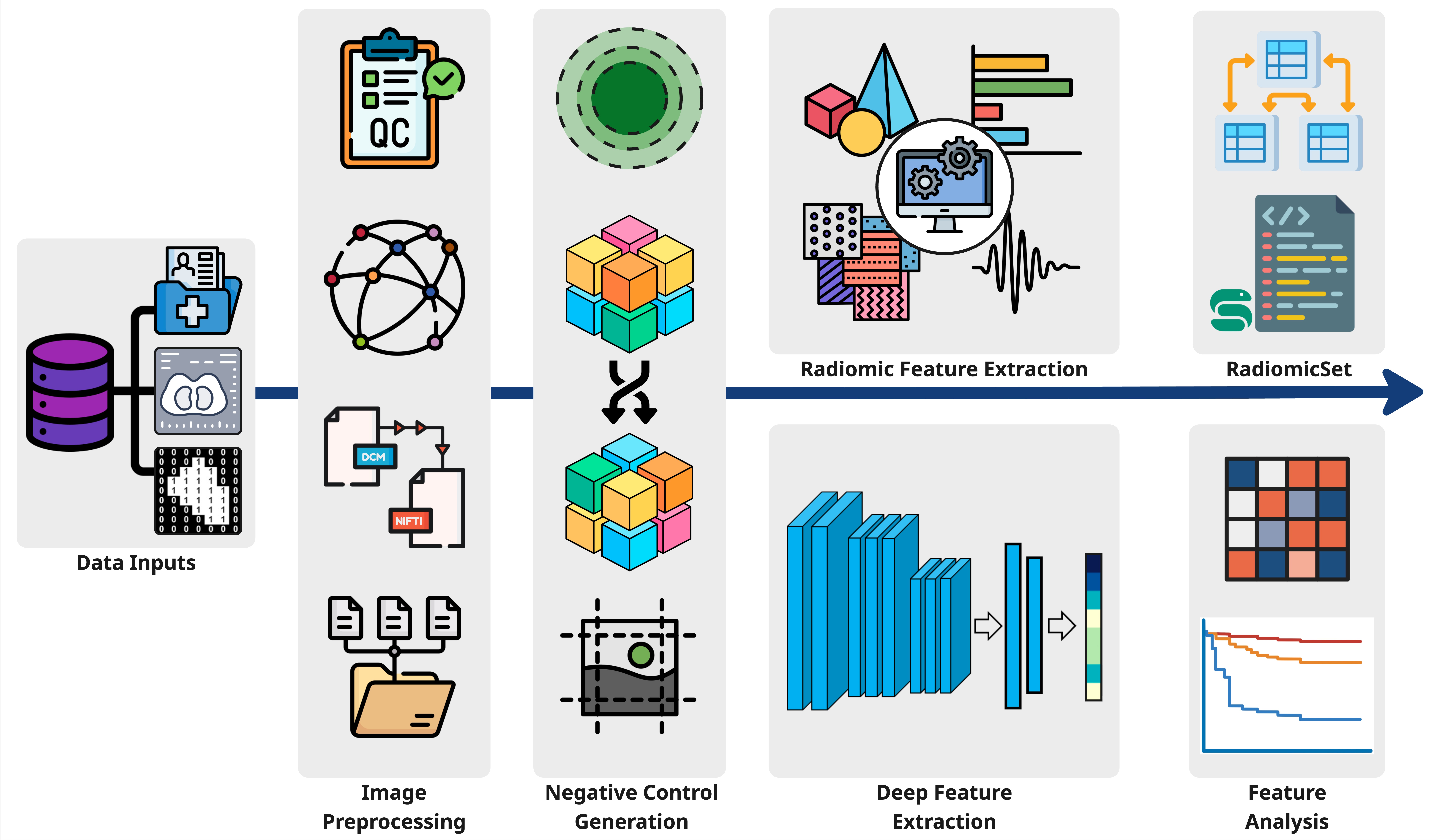}
    \caption{
    Overview of the READII-2-ROQC pipeline. Data inputs comprise DICOM images, segmentation masks, and associated clinical data. Med-ImageTools pre-processing applies quality checks, interlaces images based on metadata, converts 2D DICOM slices to 3D NIfTI volumes, and organizes files by sample. Negative control generation is customized by region, permutation, and crop methods. Feature extraction can produce shape, intensity-based, textural, and higher-order or deep features. Downstream analysis can investigate feature correlations, prediction modelling, etc. For data sharing and reproducibility, a RadiomicSet is constructed from patient metadata, extracted features, and pipeline settings.
    }
    \label{fig:pipeline}
\end{figure}

READII‑2‑ROQC is designed to systematically evaluate the robustness and informativeness of radiomic and deep learning features and signatures by comparing features extracted from original medical images with those derived from volume‑invariant negative control images (Fig. \ref{fig:pipeline}). By integrating standardized feature extraction with multiple types of negative controls, the pipeline enables assessment of feature dependence on tumour volume, image intensity distribution, and spatial structure. In this section, we showcase READII‑2‑ROQC through large‑scale feature extraction, negative control generation, and public dissemination of harmonized radiomic datasets across multiple cancer imaging cohorts.

\subsection{Dataset Processing}
To ensure data were harmonized for the downstream analyses, we applied the READII-2-ROQC pipeline to three publicly available medical imaging datasets, comprising one lung cancer cohort and two head and neck cancer (HNC) cohorts. These datasets were processed to support radiomic signature validation and comparison against volume‑invariant negative controls.

In total, 421 gross tumour volumes (GTVs) from LUNG1, 137 from HN1, and 2,994 from RADCURE were successfully processed. A summary of the demographic and clinical characteristics of each cohort is provided in Supplementary Table 1.

\subsection{Negative Control Image Generation}
To prepare for feature extraction, nine negative control images were generated for each image-mask pair to selectively disrupt spatial structure while preserving tumour volume.
Specifically, Shuffle, Sample, and Random stochastic permutations were applied within three spatial contexts: the Full image, the Region of Interest (ROI), and the Background (Supplementary Fig \ref{fig:nsclc}). This design yielded a standardized set of nine negative control images that isolate different sources of radiomic signal.

\subsection{Feature Extraction}
To reproduce the selected published signatures, radiomic features were extracted using PyRadiomics with a standardized configuration. All default PyRadiomics features were extracted, along with the Compactness1 shape feature to enable reproduction of the Aerts et al. signature \cite{Aerts2014-og}. Features were computed from original and derived image types, including wavelet, square, square‑root, logarithm, exponential, and gradient representations, resulting in 1,317 features per image. Histogram bin width and gray‑level discretization were both set to 25, and images were resampled to isotropic voxel spacing of 1 × 1 × 1 mm³ with the linear interpolator. Shape features were computed from the full image, after which the image was cropped to the tumour mask bounding box without padding prior to extraction of intensity and texture features. 

4,096 deep learning features were extracted using the FMCIB model with publicly available pre-trained weights \cite{Pai2024-wc}. Images were cropped to 50 × 50 × 50 voxels around the centroid of the tumour mask and fed into the model without an ROI mask for feature extraction.

\subsection{Case Studies}
\subsubsection{Correlation Analysis}
To evaluate the relationships within radiomic and deep learning features, we computed Pearson correlation coefficients across features extracted from original and negative control images. Feature correlations were averaged across all GTVs for each analysis, and the absolute value was taken for visualization purposes.

\begin{figure}[ht]
    \centering
    \includegraphics[width=\textwidth]{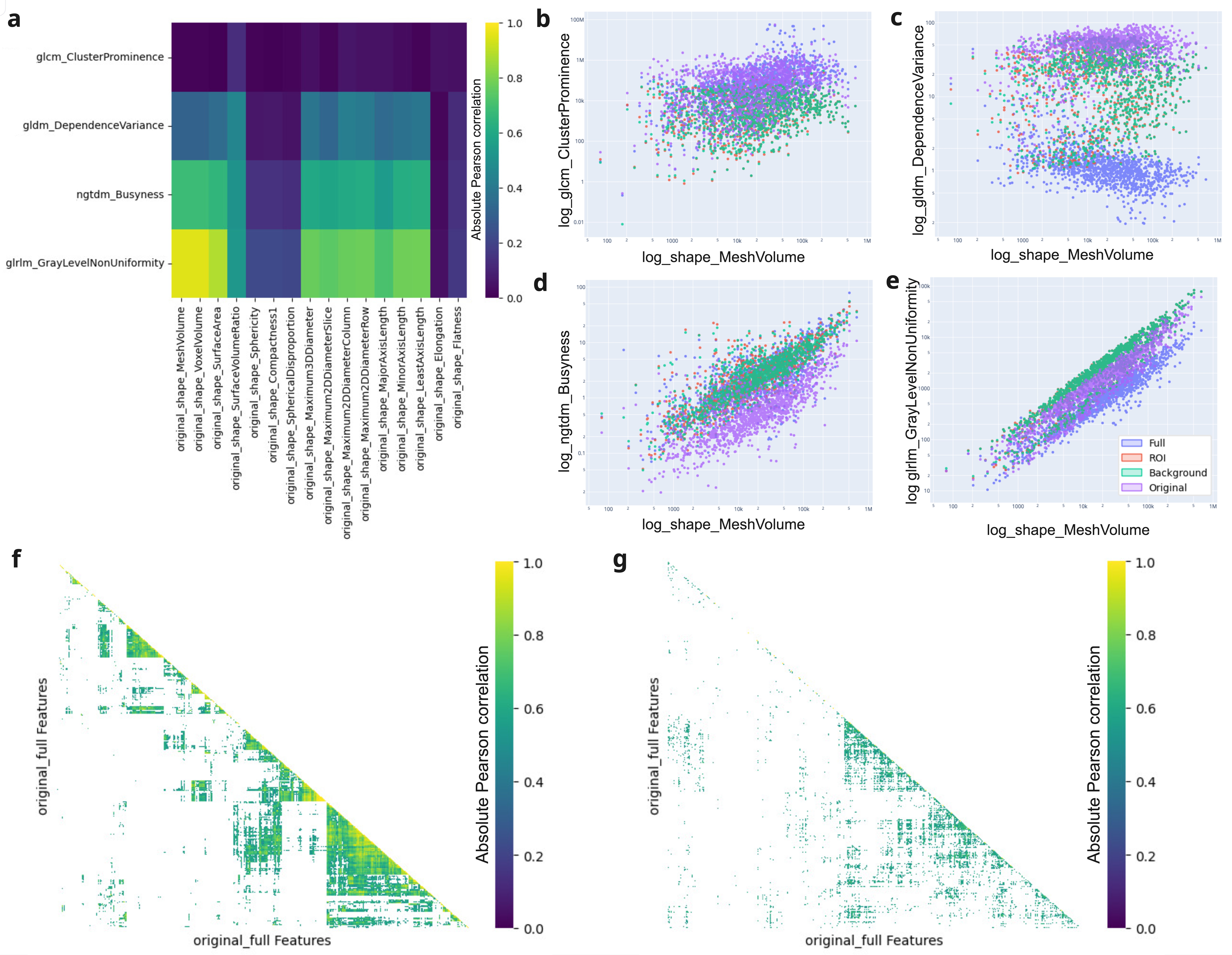}
    \caption{
    Correlation analysis results. a. Heatmap of absolute Pearson correlations averaged across all GTVs for four PyRadiomics texture features and all shape class features extracted from the ROI-Shuffle negative control images. b-e. GLCM Cluster Prominence (b), GLDM Dependence Variance (c), NGTDM Busyness (d), and GLRLM Gray Level Non-Uniformity (e) as a function of volume for the Original (purple), ROI-Random (blue), ROI-Sample (red), and ROI-Shuffle (green) Sample negative control images. f Clustermap of absolute Pearson correlations for PyRadiomics features averaged across all GTVs, clustered with Ward’s method and Euclidean distance, thresholded at 0.5. g Clustermap of absolute Pearson correlations for FMCIB features averaged across all GTVs for, clustered with Ward’s method and Euclidean distance, thresholded at 0.5.
    }
    \label{fig:velichko}
\end{figure}

Across the original images in all three datasets, strong within-class correlations were observed among radiomic shape features, as expected (Fig. \ref{fig:abscorr}a). In addition, multiple intensity and texture features demonstrated moderate-to-high correlations with shape features, indicating potential dependence on tumour volume.  These results highlight a subset of features that may be removed through conventional correlation-based filtering prior to modeling (Fig. \ref{fig:velichko}f,g). 

\begin{figure}[h]
    \centering
    \includegraphics[width=\textwidth]{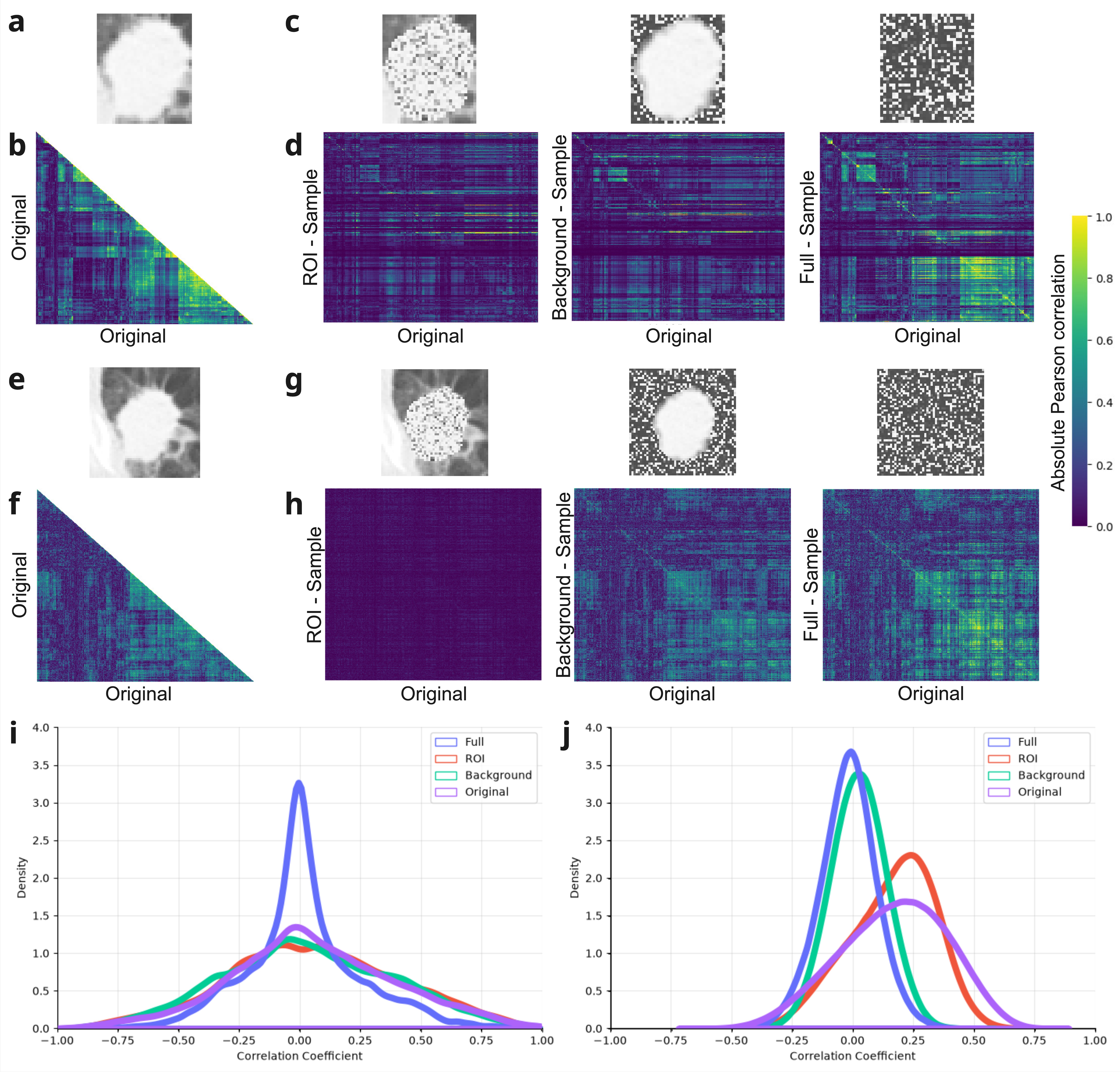}
    \caption{
    Clustered feature analysis. a. 2D slice of a bounding box cropped Original lung tumour image input to PyRadiomics. b. Clustered Pearson correlations for PyRadiomics features averaged across all GTVs. c. Lung tumour slice (a) with the Sample perturbation applied (left to right) to the ROI, Background, and Full regions. d. Clustered absolute differences in PyRadiomics feature Pearson correlation values between Original and Sample negative control images, averaged across all GTVs. e. 2D slice of a cube cropped Original lung tumour, resized to 50x50 for input to FMCIB. f. Clustered Pearson correlation values between extracted FMCIB features, averaged across all GTVs. g. Lung tumour slice (e) with the Sample perturbation applied (left to right) to the ROI, Background, and Full regions. h. Clustered absolute differences in FMCIB feature Pearson correlation values between Original and Sample negative control images, averaged across all GTVs. i. Kernel density estimate plot of PyRadiomic feature correlation distributions from the Original, Full-Sample, ROI-Sample, and Background-Sample images. j Kernel density estimate plot of FMCIB feature correlation distributions from the Original, Full-Sample, ROI-Sample, and Background-Sample images.
    }
    \label{fig:cluster}
\end{figure}

To assess whether such filtering is sufficient to eliminate volume dependence, we compared feature correlations between original images and Sample negative control images (Fig. \ref{fig:cluster}). The difference heatmaps (Fig. \ref{fig:cluster}d) and kernel density estimation plot (Fig. \ref{fig:cluster}i) highlight how the PyRadiomic feature correlations are impacted the most when the Full region is permuted, but remain intact when the ROI or Background is altered. This indicates that many radiomic features are insensitive to the loss of spatial structure and may primarily encode intensity distribution or geometric information rather than true texture. 

When extending this analysis to the FMCIB feature set, a distinct pattern emerged. Features extracted from ROI-based negative control images showed negligible differences in correlation structure relative to the original images (Fig. \ref{fig:cluster}h). However, when perturbations were introduced to the Background, the correlations diminished, regardless of if the ROI was perturbed (Fig. \ref{fig:cluster}j). This suggests that these features are insensitive to perturbations within the tumour volume and are instead primarily derived from information at the tumour boundary or from surrounding background regions. As such, these features may not capture intrinsic intratumoural heterogeneity, but rather encode contextual or geometric properties external to the ROI.

To investigate the effect of perturbation on individual radiomic feature behaviour, we examined four representative texture features with increasing levels of correlation with tumour volume (Fig. \ref{fig:velichko}a-e). Features with stronger volume dependence exhibited consistent relationships with tumour volume across all negative control configurations, with minimal deviation between original and permuted images, as showcased by GLRLM Gray Level Non-Uniformity (Fig. \ref{fig:velichko}d), one of the features included in the Aerts et al. overall survival signature \cite{Aerts2014-og}. Conversely, some features with weaker volume correlation, like GLCM Cluster Prominence (Fig. \ref{fig:velichko}b), remained largely unchanged under spatial perturbation, suggesting limited biological or structural relevance despite apparent independence from volume. Finally, features that were sensitive to negative control perturbations while maintaining lower correlation with volume, such as GLDM Dependence Variance (Fig. \ref{fig:velichko}c), represent candidates with potential biological interpretability. 
 
Together, these results demonstrate that correlation filtering alone is insufficient to ensure volume-independent radiomic features and underscore the utility of negative control-based analysis for identifying features that genuinely capture spatially meaningful information. 

\subsubsection{Signature Reproduction}
We reproduced three previously published radiomic signatures using the READII-2-ROQC pipeline and evaluated their performance on both original and negative control images, alongside baseline prediction models using tumour volume alone.

\begin{figure}[ht]
    \centering
    \includegraphics[width=\textwidth]{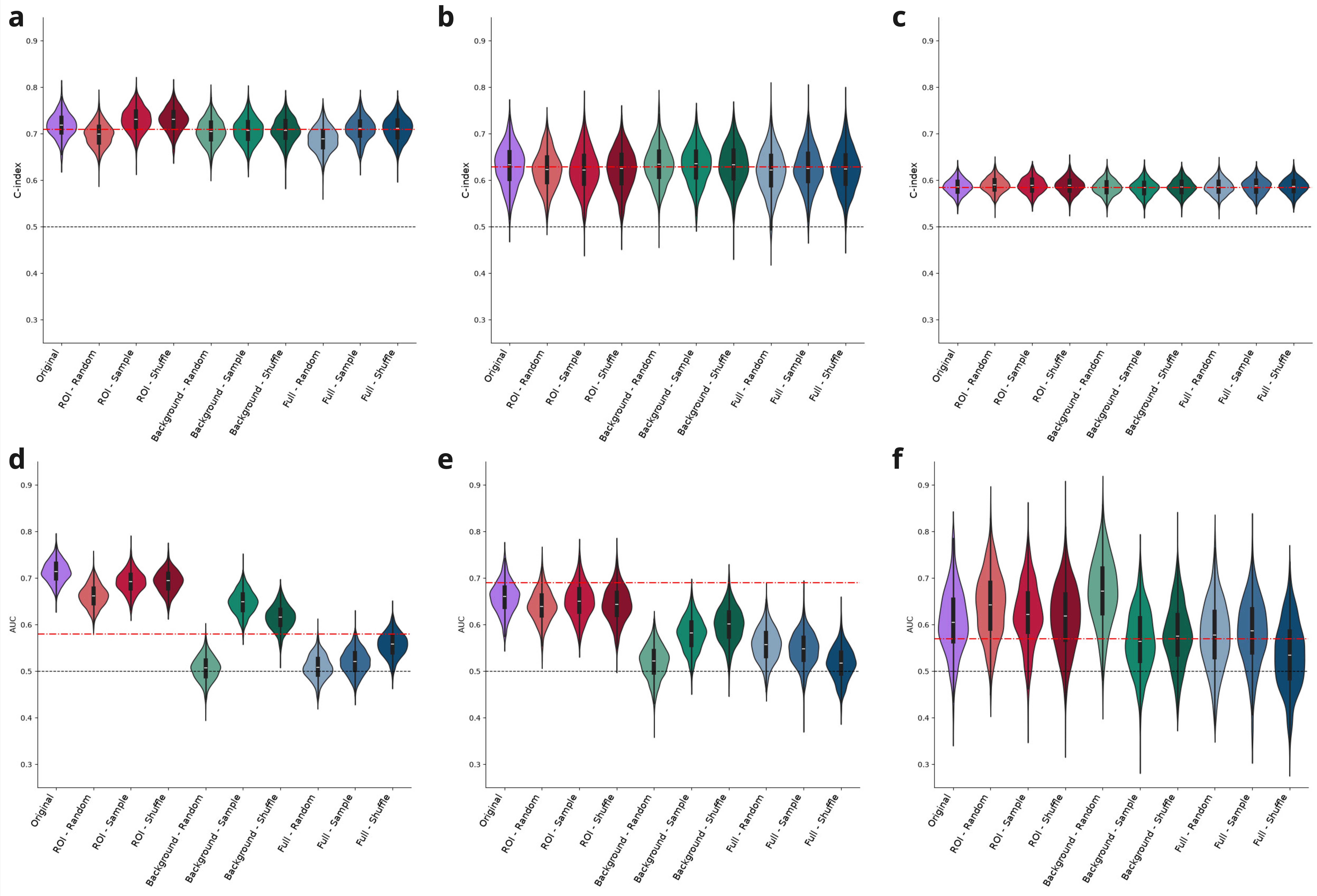}
    \caption{
    Violin plots for evaluation of radiomic signatures with READII-2-ROQC negative control images. Predictions and 95\% CIs were generated with 1000-times stratified bootstrapping for each image type. Black dashed line at 0.50 (random). Red dash-dot line shows prediction result for the Mesh Volume feature for each dataset. a-c. Overall survival prediction C-index results with the Aerts et al. radiomic signature applied to the RADCURE (a), HN1 (b) and LUNG1 (c) datasets. d-e. HPV status prediction Area Under the Curve (AUC) results with the Choi et al. radiomic signature used in a logistic regression model. The model was trained with OPC patients from the RADCURE training set (d), tested with the RADCURE test set (e), and externally validated with the HN1 dataset (f).
    }
    \label{fig:violin}
\end{figure}

The four-feature survival signature proposed by Aerts et al. \cite{Aerts2014-og} demonstrated comparable predictive performance to a volume-only model across all datasets (Fig. \ref{fig:violin}a-c). Importantly, its performance remained stable across all negative control configurations, including those that disrupted spatial texture while preserving tumour volume. These findings confirm that the predictive capacity of this signature is largely driven by volume-related information rather than independent texture signal, consistent with prior observations.	

The Choi et al. \cite{Choi2020-gz} survival signature exhibited lower predictive performance than the volume-only baseline in our analysis (Fig. \ref{fig:choi}). Furthermore, its performance did not improve when applied to original images relative to negative control images, suggesting limited robustness and potential overfitting of the signature to the original dataset. While discrepancies with the original study may reflect differences in pre-processing, feature extraction, or model specification, the lack of improvement over volume indicates limited added value of the radiomic features in this setting. 

In contrast, the Choi et al. \cite{Choi2020-gz} HPV status signature demonstrated improved performance over the volume-only model when applied to original images (Fig. \ref{fig:violin}d-f). However, its performance was differentially affected by negative control perturbations depending on the spatial region modified. Specifically, perturbations applied to the Full and Background regions resulted in substantial performance degradation, whereas perturbations confined to the ROI had a comparatively smaller effect. 

This pattern suggests that the predictive signal of the HPV signature is not derived solely from intratumoural features, but also from contextual information at the tumour boundary or in peritumoural regions. These findings highlight the potential importance of tumour–stroma interactions and support further investigation using spatially localized negative control strategies, such as contraction and expansion ring regions (Supplementary Fig. \ref{fig:regions}e-i), to more precisely characterize the origin of predictive signal.

\section{Discussion}
\label{sec:discussion}
Radiomic feature extraction presents as a non-invasive, low-cost alternative for cancer characterization, prognosis, and treatment stratification. However, clinical implementations of radiomics are barred by inconsistent and unclear methodology, data scarcity for validation, and confounding volume dependence, all contributing to irreproducibility \cite{Kocak2025-np, Lambin2025-sr, Whybra2024-hr}. In this study, we introduced the READII-2-ROQC pipeline as a standardized framework for radiomic processing and quality control, with a specific focus on disentangling biologically meaningful signal from geometric and statistical artifacts.

The negative control framework implemented in READII-2-ROQC provides a systematic approach to address this limitation. By generating voxel-permuted images that preserve tumour volume while selectively disrupting spatial structure, the pipeline enables direct assessment of whether features and signatures retain predictive performance in the absence of meaningful texture information. Across our case studies, this approach proved effective in distinguishing between volume-driven and potentially biologically informative signals.

Through large-scale application across three independent imaging cohorts, we demonstrate that many commonly used radiomic features, and by extension, published radiomic signatures, remain strongly influenced by tumour volume. Our image perturbation analysis revealed widespread dependencies between shape features and both intensity and texture features, consistent with prior observations that many radiomic descriptors act as proxies for tumour size rather than independent biomarkers \cite{Ger2019-nb, Traverso2020-zx, Velichko2021-oq, Welch2019-fv}. Importantly, we show that conventional approaches such as correlation filtering are insufficient to eliminate these effects, as numerous features remain stable under spatial perturbations that disrupt true texture signal. The analysis of FMCIB features further highlights that certain feature classes may derive signal predominantly from tumour boundaries or background regions rather than the tumour itself. This behaviour aligns with other perturbation-based explainability methods \cite{Fong2019-lz, Fong2017-xo, Ivanovs2021-mv, Petsiuk2018-vo}. By combining correlation analysis with structured negative control experiments, READII-2-ROQC enables more rigorous assessment of feature validity and supports the development of radiomic models that capture true biological signal beyond geometric confounders.

Consistent with prior work \cite{Welch2019-fv}, we confirm the survival signature proposed by Aerts et al. \cite{Aerts2014-og} does not outperform a simple volume-based model and remains robust to all negative control perturbations, indicating that its predictive power is largely volume-derived. We extended this finding to the RADCURE cohort, demonstrating the persistence of this limitation across datasets. In contrast, the Choi et al. \cite{Choi2020-gz} survival signature exhibited lower performance than volume alone in our implementation, suggesting limited robustness and highlighting the sensitivity of radiomic models to implementation details such as pre-processing, feature extraction settings, and cohort composition.

More encouragingly, the Choi et al. HPV status signature demonstrated improved predictive performance relative to volume on the original images and exhibited sensitivity to certain negative control perturbations. Notably, performance degradation was most pronounced when perturbations were applied to the full image or background regions, with comparatively smaller effects observed for ROI-restricted perturbations. This pattern suggests that the predictive signal may arise, at least in part, from tumour boundary or peritumoural characteristics rather than strictly intratumoural features. These findings align with emerging evidence supporting the relevance of tumour-stroma interactions \cite{Fan2023-uk} and peritumoural radiomics \cite{Sun2018-wo, Zhang2025-yc} for biologically meaningful prediction tasks.

Taken together, our results reinforce a critical distinction in radiomics: not all features that appear predictive or prognostic are biologically informative. Features that are invariant to spatial perturbation yet correlated with volume are unlikely to reflect meaningful tumour heterogeneity. Conversely, features that demonstrate sensitivity to structured perturbations while remaining independent of volume represent stronger candidates for biologically grounded biomarkers. The READII-2-ROQC framework provides a practical and scalable mechanism to identify these features prior to model development.

Despite these advances, several limitations remain. Variability introduced upstream of radiomic analysis, including differences in imaging acquisition, reconstruction parameters, and segmentation, continues to pose a major challenge \cite{Horvat2024-la, Reiazi2021-kb}. While harmonization strategies such as ComBat \cite{Carre2022-lx, Da-Ano2020-hj, Johnson2007-kt, Leithner2023-kd} and deep learning-based approaches are actively being explored, they were not explicitly addressed in this study. Additionally, while our negative control framework evaluates spatial and intensity dependencies, further extensions could incorporate more complex perturbations, including modality-specific transformations and biologically informed simulations.

Future work will focus on expanding the READII framework to incorporate peritumoural region analysis and integration with radiogenomic datasets. The inclusion of region-specific negative control strategies, such as contraction and expansion rings, will enable finer localization of predictive signal and further support interpretation within tumour microenvironment contexts. Alignment with emerging reporting standards \cite{Kocak2023-yt, Kocak2024-ae} will also be prioritized to enhance transparency and reproducibility.

In conclusion, READII-2-ROQC provides a robust, modular framework for evaluating the validity of radiomic features and signatures. By explicitly testing the dependence of predictive performance on tumour volume and spatial structure, the pipeline addresses a central barrier to clinical translation and supports the development of more interpretable, biologically meaningful radiomic biomarkers.

\section{Methods}
\label{sec:methods}
\subsection{Pipeline}
\subsubsection{Image Pre-processing and Standardization}
Initial cohort processing is performed using the autopipeline tool from Med-ImageTools \cite{Kim2024-xg}, which automates the conversion of raw DICOM images and associated segmentation masks into the NIfTI format. The pipeline extracts critical DICOM header metadata to ensure spatial registration between images and masks, generating a unique sample identifier for each image-mask pair to maintain full data provenance. For masks with multiple ROIs (e.g. multiple tumours), the pipeline programmatically separates distinct ROIs into individual files. A comprehensive metadata index, including original and converted image filepaths, is generated for downstream processing.

\subsubsection{Negative Control Generation Framework}
\begin{figure}[ht]
    \centering
    \includegraphics[width=\textwidth]{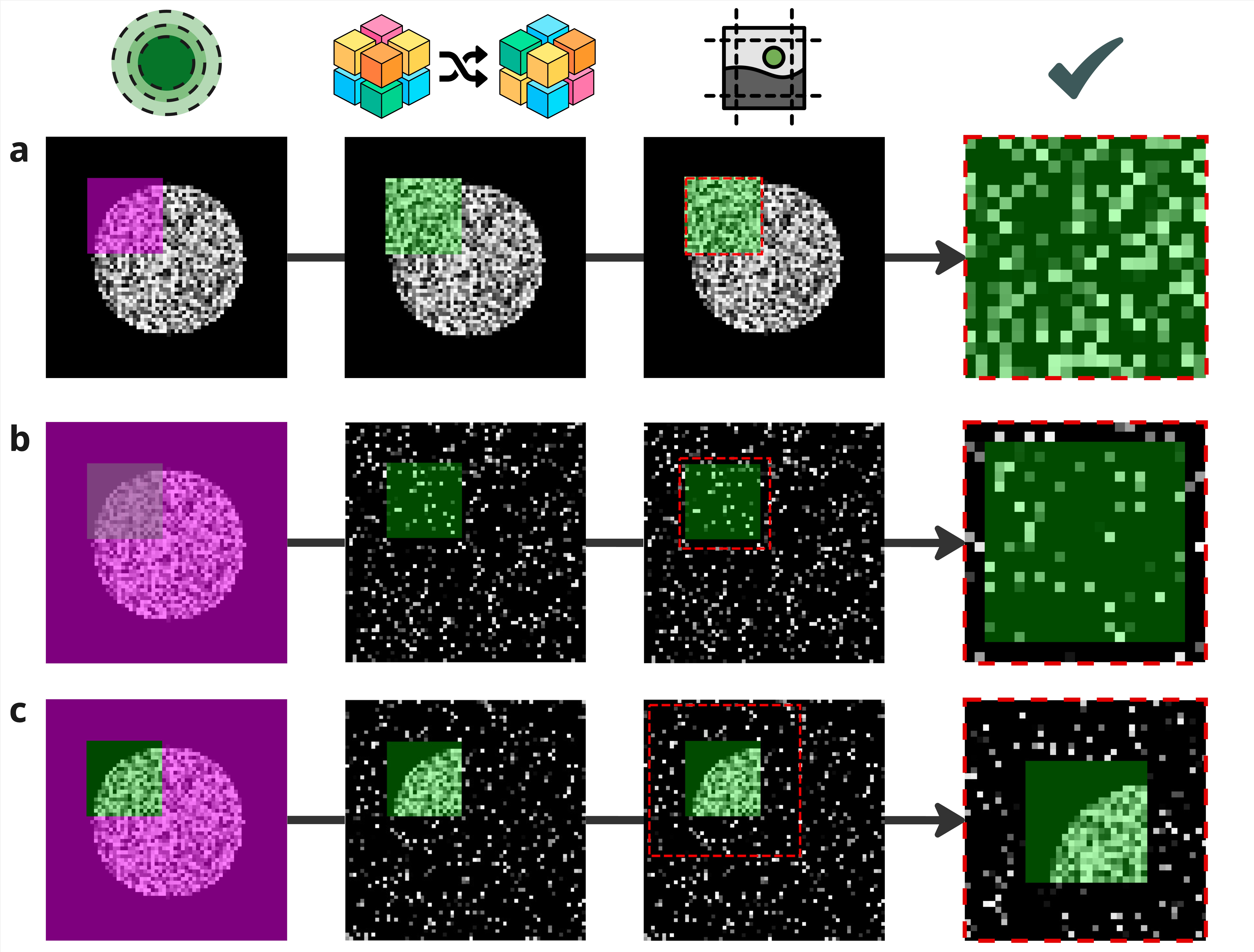}
    \caption{
    READII-2-ROQC negative control example combinations. Each image undergoes region selection; permutation of the voxels in that region; and cropping. The green square represents a region of interest (ROI) mask. Magenta sections indicate the region of the image the negative control would be applied to. A. The Random permutation applied to the ROI region and cropped with the bounding box method. B. The Sample permutation applied to the Full region and cropped with the cube method. C. The Shuffle permutation applied to the Background region and cropped with the centroid method with a dimension of 40 voxels.
    }
    \label{fig:combos}
\end{figure}

To systematically decouple image signals from geometric attributes, we developed a modular framework for generating negative control images. This framework extends the voxel randomization concept \cite{Welch2019-fv} by introducing customizable combinations of spatial regions and stochastic permutations (Fig. \ref{fig:combos}).

Permutation strategies include: (i) Random: voxel intensities are drawn from a discrete uniform distribution based on the region's original intensity range (Fig. \ref{fig:combos}a); (ii) Sample: voxel values are selected via resampling with replacement from the original distribution, maintaining the statistical frequency of intensities while introducing stochastic noise (Fig. \ref{fig:combos}b); and (iii) Shuffle: existing voxel values are randomly reordered within the region, preserving the exact intensity histogram but destroying all spatial texture (Fig. \ref{fig:combos}c).

Permutations can be localized to specific spatial regions derived from the ROI mask, including: the ROI itself (Fig. \ref{fig:combos}a); the full image (Fig. \ref{fig:combos}b); the background (extratumoural voxels) (Fig. \ref{fig:combos}c); contraction/expansion zones (core or peri-tumoural voxels) (Supplementary Fig. \ref{fig:regions}e-h); and bi-directional rings to isolate the tumour-stroma interface (Supplementary Fig. \ref{fig:regions}i). Dimensions for the contraction, expansion, and bi-directional regions can be set to a specific size (e.g. millimeters) or calculated based on the ROI volume (e.g. as a percentage).

The direction, origin, and spacing of the output negative control image is copied from the original image to maintain mask registration. As an example, to generate the negative control from Welch et al., the ROI region and Shuffle permutation would be combined. 

To ensure compatibility with deep learning architectures and traditional feature extractors, the pipeline includes three customizable cropping strategies: bounding box, cube, and centroid. For the first, a bounding box with dimensions equal to the boundaries of the ROI mask is generated (Fig. \ref{fig:combos}a). The cube method takes the ROI derived bounding box and expands it to a cube based on the largest of the mask’s three dimensions (Fig. \ref{fig:combos}b). If resizing is required, image coordinates are resampled via SimpleITK linear (images) or nearest-neighbour (masks) interpolation. In centroid, a bounding box is centered on the ROI mask centroid and expanded to a cube with user-defined dimensions or a minimum size as required by the SimpleITK library (Fig. \ref{fig:combos}c). Note that cropping and resizing are performed after negative control generation so as not to impact the distribution of available voxel intensities.

\subsubsection{Feature Extraction and Quality Control}
Two feature extraction methods were utilized in this analysis: PyRadiomics \cite{van-Griethuysen2017-wn} for radiomic features and the Foundation Model for Cancer Imaging Biomarkers (FMCIB) \cite{Pai2024-wc} for deep features. PyRadiomics (version 3.0.1a) takes an image and binary ROI mask of any dimension as input, extracts shape features, then crops the image to the boundaries of the mask with no padding prior to extracting the remaining features with any specified filter classes. The FMCIB expects a 50 × 50 × 50 voxel cropped image volume centered on the ROI. In our pipeline, we apply the crop methods described above to generate the images for deep feature extraction. No further processing is applied by the FMCIB algorithm. 

The pipeline utilizes an extraction index to orchestrate feature calculation across original and negative control datasets. This index is configured for tool-specific requirements and incorporates a final validation check of the image geometry (origin, direction, and spacing). Features are extracted based on the READII-2-ROQC, PyRadiomic, and FMCIB extraction configuration files.

\subsubsection{Radiomicset Object}
To set up the radiomic feature data to be shared, we followed the transparent approach used in our ORCESTRA cloud-based platform designed to share large multimodal biomedical data \cite{Mammoliti2021-tg}, to ensure our radiomic datasets meet the highest standards of reproducibility. 
	
To achieve this, each dataset is processed via a Snakemake \cite{Molder2021-fb} pipeline, ensuring all steps from raw DICOM to final features are version-controlled and reproducible. Data is encapsulated in our custom RadiomicSet MultiAssayExperiment R object \cite{Ramos2017-os} (Supplementary Fig. \ref{fig:radiomicset}), with each combination of feature extraction protocol and image type (original or negative control) stored as an ExperimentList. Each SummarizedExperiment is composed of: patient metadata captured in colData; feature extraction configuration settings stored as metadata; and image features stored in a BumpyMatrix, with each index representing a unique combination of the image modality and ROI masks available for that patient. Completed datasets are assigned a unique DOI via Zenodo, and the ORCESTRA API automatically generates a "Data Nutrition Label," which summarizes the pipeline version, metadata, and provenance for end-users.

\subsection{Case Studies}
\subsubsection{Datasets}
DICOM images, segmentation masks, and associated clinical data were obtained from The Cancer Imaging Archive (TCIA) \cite{Clark2013-er} using the National Biomedical Imaging Archive API. 

The NSCLC-Radiomics dataset (LUNG1) \cite{Aerts2019-rx} comprises CT images, manually delineated GTV segmentations produced by a radiation oncologist, and clinical outcome data for 422 non-small cell lung cancer (NSCLC) patients. This dataset was used in the original radiomics study by Aerts et al. \cite{Aerts2014-og}.

The Head-Neck-Radiomics-HN1 (HN1) \cite{Wee2019-ye} includes CT images, GTV segmentations and clinical data from 137 head and neck cancer (HNC) patients treated with radiotherapy. This dataset was used in the original study by Aerts et al. \cite{Aerts2014-og} This data was accessed under the TCIA No Commercial Limited Access License. 

The RADCURE dataset \cite{Welch2023-od} consists of 3,346 HNC CT image volumes, corresponding radiotherapy structure sets (RTSTRUCT) containing primary GTV contours, and detailed clinical data. This data was accessed under the TCIA Restricted License and was used for external validation where applicable.

\subsubsection{Correlation Analysis}
To characterize dependencies among imaging features, we computed Pearson correlation coefficients for radiomic and deep learning features extracted from original images and their matched negative control images. Correlation analyses were performed across corresponding original-negative control feature sets to assess whether feature relationships were preserved or disrupted by spatial perturbation. Impacts to feature correlation were captured by calculating the absolute difference between intra-correlations of the original image features and inter-correlations of original and negative control image features. Thresholding of the differences was investigated as a method of identifying and filtering volume-invariant features.

Correlation matrices were visualized as heatmaps. PyRadiomic features were organized by feature class. FMCIB deep features do not have defined classes, so hierarchical clustering was applied with the average Ward distance to aid visualization of correlations. 

We further investigated individual radiomic feature behaviour by examining representative texture features spanning a range of correlations with shape-derived metrics. We adapted plots presented in Velichko et al. \cite{Velichko2021-oq} to illustrate the relationship between volume dependence and the negative control perturbations.

Kernel density estimate plots were generated to assess the effect of the perturbations on the distribution of all feature correlations for PyRadiomics and FMCIB.

\subsubsection{Signature Reproduction}
Three previously published radiomic signatures were reproduced using the READII-2-ROQC pipeline. The first signature, originally proposed by Aerts et al. \cite{Aerts2014-og} and subsequently reproduced by Welch et al. \cite{Welch2019-fv}, aimed to predict overall survival in lung and HNC patients, and was developed and validated with the LUNG1 and HN1 datasets, respectively. The signature is defined as:
\begin{multline}
(1.74e^{-11} * Original\_FirstOrder\_Energy) + (-1.65e^{01} * Original\_Shape\_Compactness1) \\
+ (4.95e^{-05} * Original\_GLRLM\_GrayLevelNonUniformity) \\
+ (2.81e^{-06} * Wavelet\_HLH\_GLRLM\_GrayLevelNonUniformity)
\end{multline}
This signature was applied to the original images and negative control images generated from the same datasets, as well as to the RADCURE dataset as an external validation cohort.
	
The second signature, reported by Choi et al. \cite{Choi2020-gz}, predicts survival in oropharyngeal cancer (OPC) patients and was validated using the OPC subset of HN1. This signature is defined as: 
\begin{multline}
(0.32 * Original\_Shape\_SphericalDisproportion) + (-2.363*10^{-03} * Original\_FirstOrder\_Minimum) \\
+ (-1.753*10^{-05} * Original\_FirstOrder\_10Percentile)
\end{multline}
This signature was applied to the same OPC subset of HN1 and to OPC patients from RADCURE, using both original and negative control images.

The third signature, also from Choi et al., predicts human papillomavirus (HPV) status in OPC patients. This signature was developed using generalized linear models on the same dataset as the survival signature. In our study, the nine radiomic features selected were used to fit a logistic regression model with the training cohort of RADCURE OPC patients. Model performance was tested with the RADCURE test cohort, and externally validated on the HN1 dataset.

\subsubsection{Model Comparison and Evaluation}
Survival and HPV status radiomic signatures were compared against a baseline Cox proportional hazards (CPH) or logistic regression model, respectively, fit with only the Mesh Volume feature. For the RADCURE cohort, the volume‑only model was trained on the designated training subset and evaluated on the testing subset. For HN1 and LUNG1, volume‑only models were trained and evaluated on the full datasets. The HPV signature was compared to a logistic regression model using

Survival model performance was assessed using Harrell’s concordance index (C‑index) \cite{Harrell1996-ah}, computed with the scikit-survival package (version 0.25.0) \cite{Polsterl2020-bz}. HPV classification performance was evaluated using area under the receiver operating characteristic curve (AUC), from the scikit-learn package. Confidence intervals were estimated via 1,000 bootstrap resamples.

\subsection{Data Availability}
All datasets used in this work are publicly available through The Cancer Imaging Archive. RADCURE \cite{Welch2023-od} (\url{https://www.cancerimagingarchive.net/collection/radcure/}), HN1 \cite{Wee2019-ye}  (\url{https://www.cancerimagingarchive.net/collection/head-neck-radiomics-hn1/}), LUNG1 \cite{Aerts2019-rx} (\url{https://www.cancerimagingarchive.net/collection/nsclc-radiomics/}). 
	
To support transparency and reuse, the complete radiomic feature sets generated by READII‑2‑ROQC have been published on ORCESTRA (\url{https://www.orcestra.ca/radiomicset}). For each processed dataset, the published resources include radiomic and deep learning features extracted from the original images and all nine negative controls, associated clinical metadata, and the exact feature extraction settings used. This enables independent validation, benchmarking of radiomic methods, and extension of READII‑2‑ROQC analyses to additional radiomic signatures and modeling frameworks.

\subsection{Code Availability}
Code used to produce results are available on GitHub. The main READII-2-ROQC repository (\url{https://github.com/bhklab/readii_2_roqc/}) is set up for image pre-processing, including running Med-ImageTools (\url{https://github.com/bhklab/med-imagetools/}), negative control generation, and radiomic feature extraction. Configuration settings used for each dataset are described in Supplementary Table 2. PyRadiomics features were extracted with the linear\_all\_images\_features.yaml settings file, available in the READII-2-ROQC repository. For deep feature extraction with FMCIB, a separate repository is available (\url{https://github.com/bhklab/readii-fmcib/}) to demonstrate how to run READII-2-ROQC with an external feature extraction method and manage dependency conflicts.

\newpage
\bibliographystyle{unsrt}  

\newpage

\appendix
\renewcommand{\tablename}{Supplementary Table}
\setcounter{table}{0}

\begin{table}[!ht]
 \caption{Baseline demographic and clinical characteristics of the study population.}
   \centering
    \begin{tabular}{llccc}
      \toprule
        & & \multicolumn{3}{c}{\textbf{Datasets}} \\
        \multicolumn{2}{l}{\textbf{Characteristic}} & LUNG1 (N = 441) & HN1 (N = 137) & RADCURE (N = 2994) \\ \hline
        \multicolumn{2}{l}{\textbf{Age at diagnosis (y)}}  &  &  &  \\ 
            & Median (range) & 68.6 (33.7 - 91.7) & 61.0 (44 - 83) & 61.9 (15.6 - 90.0) \\
        \multicolumn{2}{l}{\textbf{Sex}} &  &  &  \\ 
            & Male & 290 & 111 & 2382 \\
            & Female & 131 & 26 & 612 \\
        \multicolumn{2}{l}{\textbf{Smoking Status}}  &  &  &  \\ 
            & Current 0 & 0 &	1007 \\
            & Former  0 & 0 &	1178 \\
            & Non-smoker 0 & 0 & 771 \\
            & Unknown & 422 & 137 & 38 \\
        \multicolumn{2}{l}{\textbf{HPV Status}}  &  &  &  \\ 
            & Positive & 0 & 23 & 1023 \\
            & Negative & 0 & 58 & 508 \\
            & Unknown & 422 & 56 & 1463 \\
        \multicolumn{2}{l}{\textbf{Treatment}}  &  &  &  \\ 
            & Radiotherapy & 196 & 100 & 1663 \\
            & Chemo-Radiation & 226 & 37 & 1261 \\
            & Other & 0 & 0 & 70 \\
        \multicolumn{2}{l}{\textbf{T Stage}} &  &  &  \\ 
            & T0 & 0 & 0 & 158 \\
            & T1 & 93 & 35 & 640 \\
            & T2 & 155 & 32 & 834 \\
            & T3 & 53 & 24 & 781 \\
            & T4 & 117 & 46 & 528 \\
            & Tis & 0 & 0 & 38 \\
            & X & 2 & 0 & 3 \\
            & Unknown & 1 & 0 & 12 \\
        \multicolumn{2}{l}{\textbf{N Stage}}  &  &  &  \\ 
            & N0 & 170 & 60 & 1014 \\
            & N1 & 23 & 16 & 299 \\
            & N2 & 141 & 58 & 1477 \\
            & N3 & 84 & 3 & 190 \\
            & N4 & 3 & 0 & 0 \\
            & X & 0 & 0 & 1 \\
            & Unknown & 0 & 0 & 13 \\
        \multicolumn{2}{l}{\textbf{Overall Stage}}  &  &  &  \\ 
            & 0 & 0 & 0 & 38 \\
            & I & 93 & 24 & 319 \\
            & II & 40 & 11 & 353 \\
            & III & 287 & 23 & 542 \\
            & IV & 0 & 79 & 1717 \\
            & X & 0 & 0 & 5 \\
            & Unknown & 1 & 0 & 20 \\
        \multicolumn{2}{l}{\textbf{No. of Deaths}} & 373 & 74 & 1007 \\
        \bottomrule
    \end{tabular}
    \caption*{\textit{HPV = Human papillomavirus, LUNG1 = NSCLC-Radiomics, HN1=HEAD-NECK-RADIOMICS-HN1. 7th edition of TMN staging system used.}}
    \label{tab:demographic}
\end{table}

\begin{landscape}
\begin{table}[!ht]
 \caption{Dataset specific configuration settings for the READII-2-ROQC pipeline.}
   \centering
    \begin{tabularx}{\linewidth}{l l X X X}
      \toprule
        & & \multicolumn{3}{c}{\textbf{Datasets}} \\
        \multicolumn{2}{l}{\textbf{Setting}} & \multicolumn{1}{c}{LUNG1} & \multicolumn{1}{c}{HN1} & \multicolumn{1}{c}{RADCURE} \\ \hline
        \multicolumn{2}{l}{\textbf{CLINICAL}} &  &  & \\
         & FILE & NSCLC-Radiomics-Lung1.clinical-version3-Oct-2019.csv & Copy of HEAD-NECK-RADIOMICS-HN1 Clinical data updated July 2020 2.csv & RADCURE\_Clinical\_v04\_20241219.xlsx \\
         & time\_label & Survival.time & overall\_survival\_in\_days & Length FU \\
         & event\_label & deadstatus.event & event\_overall\_survival & Status \\
         & convert\_to\_years & True & True & False \\
         & event\_value\_mapping & \{\} & \{\} & \{"Alive": 0, "Dead": 1\} \\
         & EXCLUSION\_VARIABLES & \{\} & \{\} & \{'Ds Site': ['Sarcoma', 'Unknown', 'Paraganglioma', 'Salivary Glands', 'Other', 'benign tumor', 'Orbit', 'Lacrimal gland', 'Skin']\} \\
         & INCLUSION\_VARIABLES & \{\} & \{'index\_tumour\_location': ['oropharynx'], 'overall\_hpv\_p16\_status': ['positive', 'negative']\}** & \{'Ds Site': ['Oropharynx'], 'HPV': ['Yes, positive', 'Yes, Negative']\}** \\
        \multicolumn{2}{l}{\textbf{MIT}} &  &  &  \\
         & ROI\_MATCH\_MAP & GTV: GTV-1, gtv-pre-op & GTV: GTV-1 & GTV: GTVp \\
        \multicolumn{2}{l}{\textbf{ANALYSIS}} &  &  &  \\
         & split\_variable & null & null & RADCURE-challenge \\
         & train\_label & null & null & training \\
         & test\_label & null & null & test \\
         & impute & null & null & training \\
      \bottomrule
    \end{tabularx}
  \caption*{\textit{** Settings used for the HPV prediction modelling only.}}
  \label{tab:settings}
\end{table}
\end{landscape}

\renewcommand{\figurename}{Supplementary Figure}
\setcounter{figure}{0}

\begin{figure}[h]
    \centering
    \includegraphics[width=0.6\textwidth]{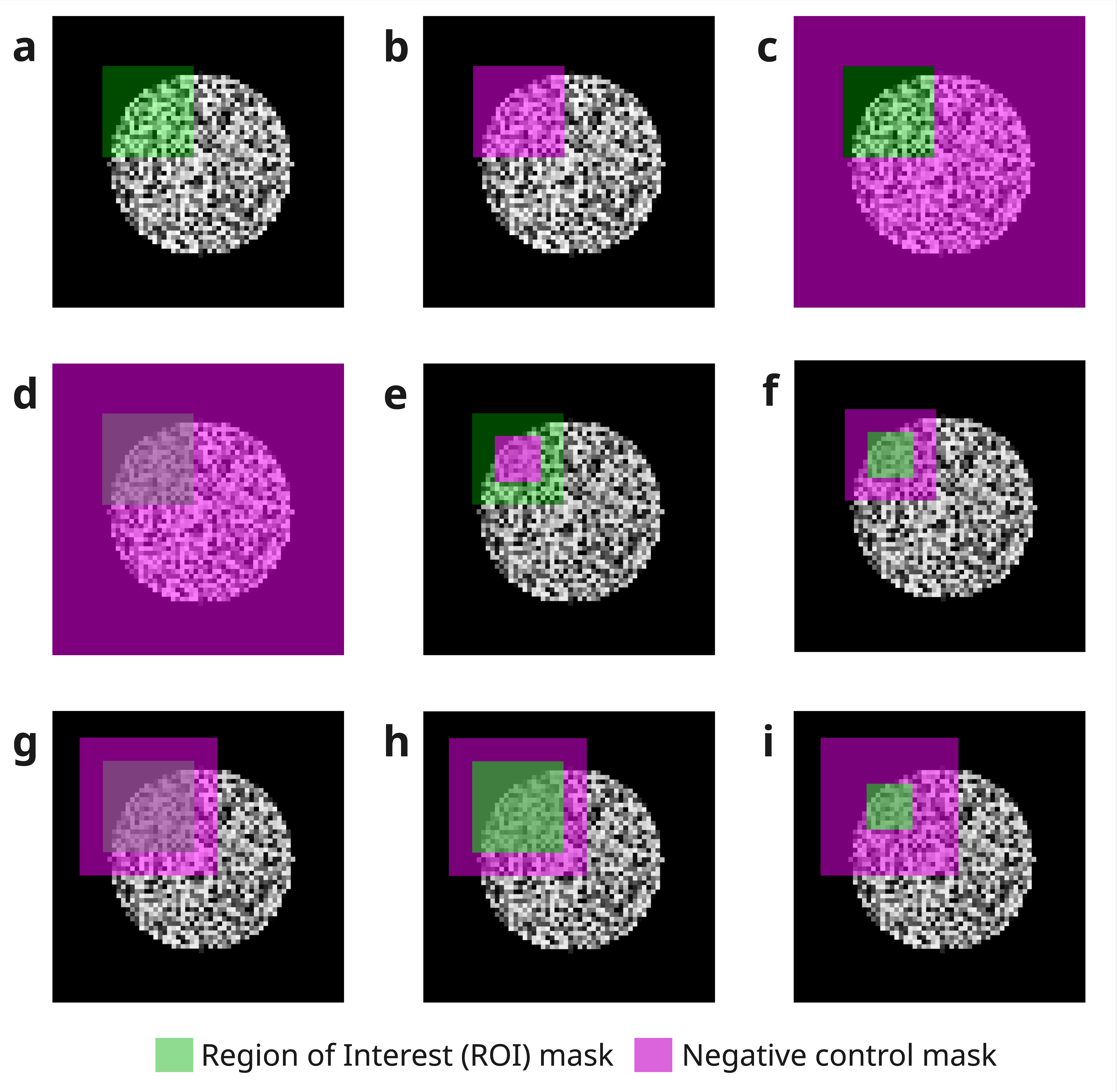}
    \caption{
    Negative control spatial region options on 2D toy images. The green square represents a region of interest (ROI) mask. Magenta sections indicate the region of the image the negative control would be applied to a. Original image; b. ROI; c. Background; d. Full; e. Contraction; f. Contraction Ring; g. Expansion; h. Expansion Ring; i. Bi-directional Ring.
    }
    \label{fig:regions}
\end{figure}

\begin{figure}[h]
    \centering
    \includegraphics[width=0.75\textwidth]{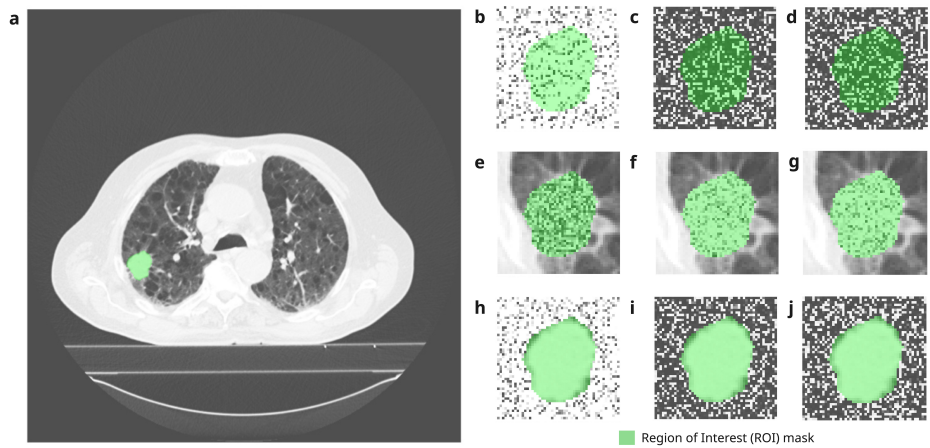}
    \caption{
    Negative control permutation and region combinations on a sample from NSCLC-Radiomics (LUNG1). a. Slice of an original lung CT with the segmented tumour mask overlaid in green. b-j Cube crop of the slice in a with the following negative control combinations: b. Full-Random; c. Full-Sample; d. Random-Sample; e. ROI-Random; f. ROI-Sample; g. ROI-Shuffle; h. Background-Random; i. Background-Sample; j. Background-Shuffle.
    }
    \label{fig:nsclc}
\end{figure}

\begin{figure}[h]
    \centering
    \includegraphics[width=0.6\textwidth]{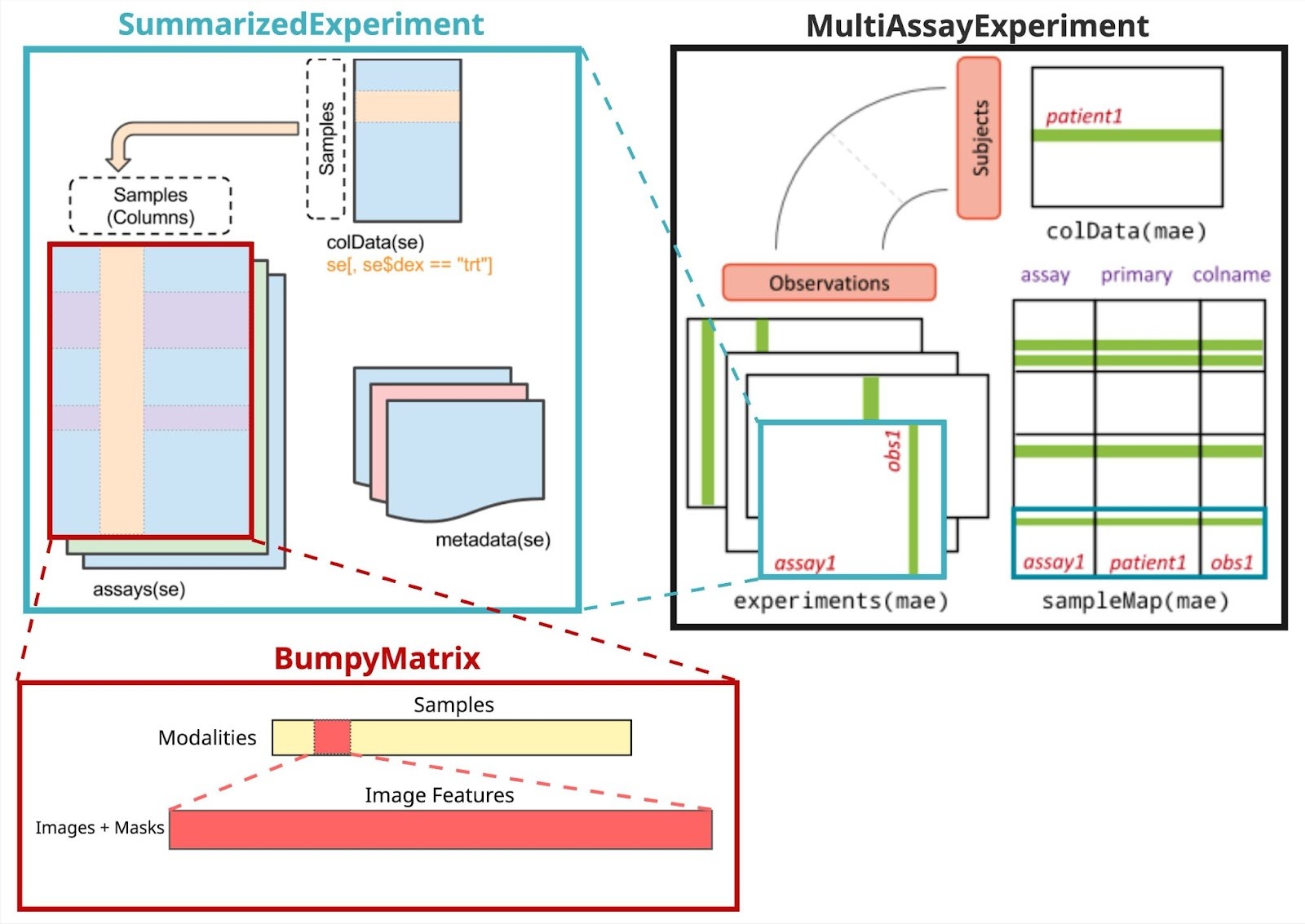}
    \caption{
    RadiomicSet ORCESTRA object. Each dataset on ORCESTRA has a MultiAssayExperiment, composed of colData containing clinical data for the patient set and a SummarizedExperiment for each extraction protocol and image type (original or negative control) combination. SummarizedExperiments contain a single BumpyMatrix assay, where each index of the matrix contains a dataframe of image features for each image and mask combination for that patient.
    }
    \label{fig:radiomicset}
\end{figure}

\begin{figure}[h]
    \centering
    \includegraphics[width=\textwidth]{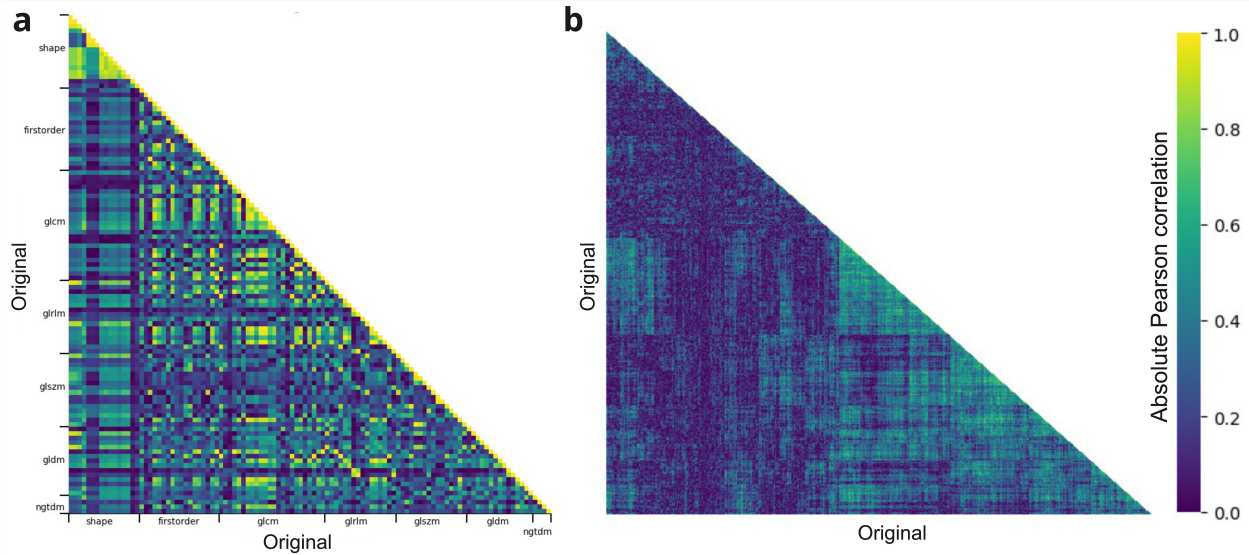}
    \caption{
    Absolute Pearson correlations of features extracted from Original images averaged across RADCURE, HN1, and LUNG1. a. Heatmap of PyRadiomic features correlations organized by feature class. b. Clustermap of FMCIB feature correlations, clustered with Ward’s method using Euclidean distance.
    }
    \label{fig:abscorr}
\end{figure}

\begin{figure}[ht]
    \centering
    \includegraphics[width=\textwidth]{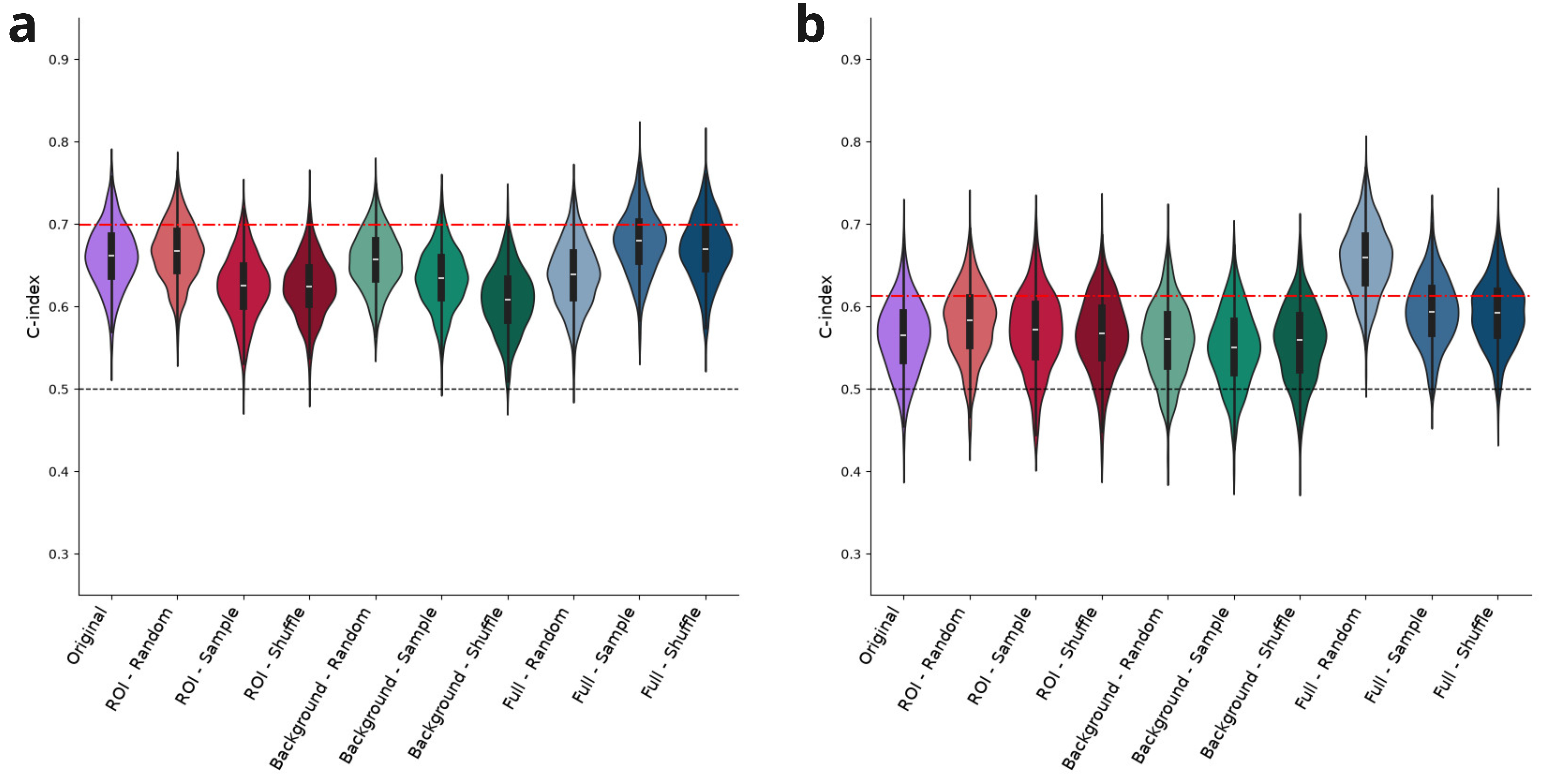}
    \caption{
    Choi et al. radiomic survival signature in oropharynx (OPC) patients READII-2-ROQC evaluation. Violin plots for survival prediction c-index results with the Choi et al. radiomic signature for OPC patients from the RADCURE (a) and HN1 (b)datasets. C-indexes and 95\% CIs were generated with 1000-times stratified bootstrapping for each image type. The black dashed line is at 0.50 (random) and the red dash-dot lines show the prediction result using Mesh Volume.
    }
    \label{fig:choi}
\end{figure}

\end{document}